\documentclass[conference]{IEEEtran}

\newcommand{\tool}{\texttt{AESOP}}
\newcommand{\approach}{\texttt{AESOP}}

\usepackage{microtype}
\usepackage{graphicx}
\usepackage{subcaption}
\usepackage{booktabs}
\usepackage{amsmath}
\usepackage{amssymb}
\usepackage{mathtools}
\usepackage{amsthm}
\usepackage{array}
\usepackage{tabularx}
\usepackage{makecell}
\usepackage{adjustbox}
\usepackage[table]{xcolor}
\usepackage{url}
\usepackage{cite}
\usepackage[most]{tcolorbox}
\usepackage{multirow}

\let\oldmultirow\multirow
\makeatletter
\renewcommand{\multirow}[1]{%
  \@ifnextchar\bgroup
    {\oldmultirow{#1}}%
    {\oldmultirow{5}{*}{#1}}%
}
\makeatother

\theoremstyle{plain}

\theoremstyle{definition}

\theoremstyle{remark}

\usepackage[textsize=tiny,disable]{todonotes}

\definecolor{ndssblue}{rgb}{0.21,0.49,0.74}
\usepackage[
    pagebackref=false,
    breaklinks=true,
    colorlinks=true,
    allcolors=ndssblue
]{hyperref}

\begin{document}

\title{AESOP: {A}dversarial {E}xecution-path {S}election to {O}verload Deep Learning {P}ipelines}

\author{
\IEEEauthorblockN{%
Tingxi Li\IEEEauthorrefmark{1},
Mingfang Ji\IEEEauthorrefmark{2},
Ravishka Shemal Rathnasuriya\IEEEauthorrefmark{1},
Simin Chen\IEEEauthorrefmark{1},
Yitao Hu\IEEEauthorrefmark{2},
Wei Yang\IEEEauthorrefmark{1}}
\IEEEauthorblockA{%
\IEEEauthorrefmark{1}The University of Texas at Dallas \\
\{tingxi.li,\,ravishka.rathnasuriya,\,simin.chen,\,wei.yang\}@utdallas.edu}
\IEEEauthorblockA{%
\IEEEauthorrefmark{2}Tianjin University \\
\{jimingfang,\,yitao\}@tju.edu.cn}
}

\maketitle

\begin{abstract}

\textcolor{black}{
Modern machine learning deployments increasingly compose specialized models into dynamic inference pipelines, where upstream components produce intermediate predictions that determine the workload and inputs of downstream components. The cost of processing an input is therefore not determined by any single model, but by two coupled factors: the per-inference cost of each invoked component and its workload volume. Because these pipelines run under hard real-time constraints, efficiency is a fundamental requirement for system availability. 
We show that this structure creates an efficiency-attack surface that existing methods targeting single models cannot exploit: on identical inputs and budgets, path-aware targeting inflates FLOPs by $2,407\times$ while the strongest single-model baseline achieves $117\times$ --- a $20\times$ gap attributable entirely to where the attack is directed. We formalize this as the adversarial path-selection problem and present \texttt{AESOP}, a framework combining vulnerability-guided path ranking with adaptive loss weighting. We evaluate \texttt{AESOP} on five pipelines plus a production-realistic deployment variant with batching, bounded buffering, and confidence-threshold defenses. \texttt{AESOP} achieves up to $2,407\times$ FLOPs and $419\times$ latency inflation in white-box setting and 58$\times$ FLOPs / 17$\times$ latency in gray-box settings. Under system-level defenses, the attack is not neutralized but redirected: pipelines are forced to choose between throughput collapse ($0.578 \to 0.006$ input/s) and $96.7\%$ data loss to sustain throughput.}
\end{abstract}

\section{Introduction \label{sec:intro}}


\textcolor{black}{While large language models (LLMs) represent a notable shift toward end-to-end architectures, most production machine learning systems --- such as video analytics, autonomous driving, content moderation, and surveillance --- are still built as multi-model pipelines.} Within these pipelines, specialized models are connected by producer-consumer dataflow, where each handles a distinct subtask and forwards its output downstream \cite{shen2019nexus, hu2021scrooge, ghafouri2023ipa, kim2023dream, zhao2026pard}. This architecture offers modular reuse of pre-trained components, flexible composition of heterogeneous pipeline topologies, and independent scaling of individual stages \cite{crankshaw2020inferlinemlpredictionpipeline}. Major inference platforms now provide first-class support for pipeline orchestration \cite{nvidia_dynamo_triton}, and recent surveys have documented the prevalence and growing importance of multi-model pipeline deployments across edge devices, cloud serving systems, and safety-critical applications \cite{Ullrich_2025, Ahmad_2024, bentoml_ai_infra_survey_2024}.

\textcolor{black}{Because these pipelines run under hard real-time constraints, latency is not merely a performance metric but a determinant of availability.} In a traffic monitoring pipeline of the kind proposed in recent multi-model designs \cite{ammar2023multi, li2025multi}, which chain specialized models for vehicle detection, license-plate recognition, and downstream identification on edge hardware, an adversarial input can stretch end-to-end inference from 45 ms to 13.6 s per frame, at which point frame dropping becomes systemic and detection coverage degrades to operational failure \cite{saurez2023utility}. In a wildlife conservation pipeline following the two-stage detection-then-classification architecture established for camera-trap monitoring \cite{mulero2025addressing, dussert2026paying}, the same class of input drives latency from 21 ms to 8.8 s and inflates per-image energy from 0.04 J to 1720 J, compressing operational autonomy by orders of magnitude \cite{velasco2024reliable, corva2022smart}. In an Amber Alert pipeline modeled on the missing-child face-recognition and age-estimation systems studied by Deb et al. \cite{deb2020child} and Dehghan et al. \cite{dehghan2017dager}, alert-generation latency stretches from 5 s to 25 s, eroding responsiveness in exactly the window the system is designed to operate, where speed of dissemination is the primary determinant of recovery outcomes \cite{griffin2008child}. In each case, an adversary who inflates pipeline cost defeats the system's availability guarantee --- the timely production of correct outputs --- placing efficiency squarely alongside integrity and confidentiality.

\begin{figure}
    \centering
    \includegraphics[width=1\linewidth]{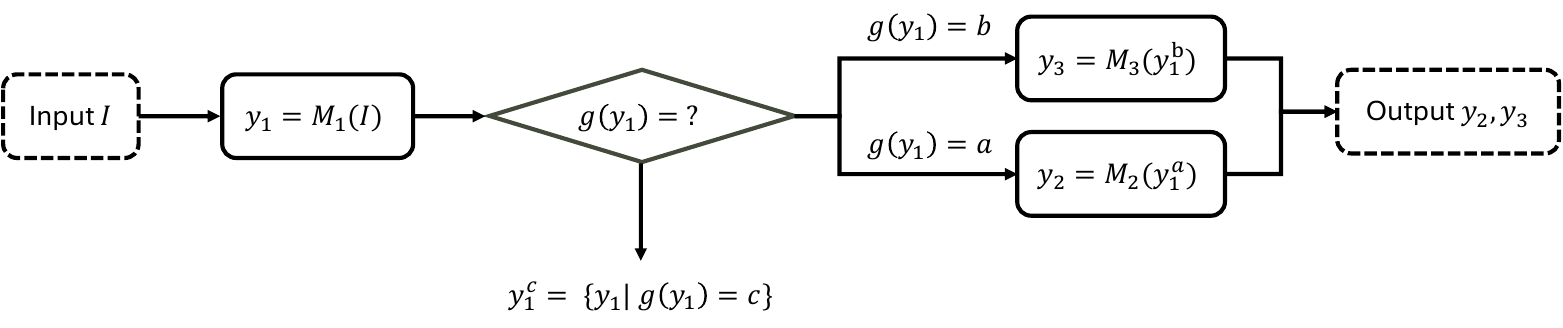}
    \caption{Execution paths in a dynamic deep learning pipeline. A gating decision $g(y_1)$ on the output of $M_1$ partitions inputs across three branches with distinct downstream costs.}
    \label{fig:execution-path}
\end{figure}

\textcolor{black}{A growing body of work studies adversarial efficiency attacks on dynamic deep learning models, whose per-inference cost and output cardinality are input-dependent — for example, early-exit networks, object detectors, and text-generation models \cite{shumailov2021sponge, shapira2023phantom, chen2024overload, ma2024slowtrack, liu2023slowlidar, chen2022nmtsloth, haque2023slothspeech, haque2020ilfo, pan2023gradmdm, pan2022gradauto}}. Across this literature, however, the attack surface is consistently a standalone model: the adversary crafts a perturbation that maximizes cost within one neural network's computation graph, whether by inflating non maximum suppression (NMS) survivors, delaying EOS tokens, or suppressing early-exit triggers. This single-model assumption matters in pipelines: the cost of processing an input is not determined by how expensive any one model becomes; it is determined by which execution path the input activates. Figure~\ref{fig:execution-path} illustrates this: a gating decision after model $M_1$ partitions output set $y_1$ into three subsets $\{y^a_1, y^b_1, y^c_1\}$,  each propagate through a distinct branch --- $y^c_1$ exits the pipeline early while $y^a_1$ and $y^b_1$ are processed by $M_2$ and $M_3$ respectively. Each branch therefore incurs a different cost, and both the per-inference cost of $M_2$ and $M_3$ and the size of each subset are themselves input-dependent and exploitable \cite{chen2024overload, chen2022nmtsloth, pan2023gradmdm}.  The pipeline's worst-case cost is bounded by its most computationally intensive path, and an adversary who can steer inputs into that path realizes the worst case at will. Existing attacks have no mechanism to reason about routing: they fix a target model in advance, which succeeds locally when that model is the only dynamic stage, and fails globally when multiple dynamic stages and execution paths coexist.  We confirm this empirically: on identical inputs and identical perturbation budgets, the strongest single-model baseline achieves 117$\times$ FLOPs amplification while a path-aware attack achieves 2,407$\times$, a 20$\times$ gap caused by the  missing algorithmic primitive: the ability to choose which path to attack.

We formalize this primitive as the adversarial path-selection problem and present \texttt{AESOP}, the first efficiency-attack framework that solves it. \texttt{AESOP} decomposes the attack into two coordinated decisions that prior work treats as one. First, vulnerability-guided path ranking profiles per-module cost, output cardinality, and gating behavior, then enumerates execution paths and ranks them by a vulnerability score that quantifies the cost amplification each path would incur under adversarial inputs. Second, adaptive loss weighting generates the perturbation against the selected path with loss terms whose weights are set by each module's contribution to the path's vulnerability score. The framework is separable from the perturbation backend: any current or future single-model efficiency attack can serve as the per-module perturbation step, with \texttt{AESOP} supplying the system-level decisions that single-model attacks structurally cannot make.

We evaluate \texttt{AESOP} on five pipelines spanning traffic monitoring, wildlife conservation, smart-home alerting, expressway surveillance, and Amber Alert generation, plus a production-realistic deployment variant of the traffic pipeline with batched inference, bounded inter-module buffering, and confidence-threshold filtering. \texttt{AESOP} achieves up to 2,407$\times$ FLOPs and 419× latency inflation in the white-box setting and 58$\times$× FLOPs and 17$\times$ latency in a gray box transfer setting.
Under deployed system-level defenses, the attack is not neutralized but redirected: confidence thresholding fails because \texttt{AESOP} produces high-confidence detections by design, and bounded buffering restores throughput only by silently dropping 96.7\% of detections — converting compute exhaustion into data loss without removing the underlying vulnerability. This tradeoff is not an artifact of our specific defenses but a structural consequence of where the attack operates: at the routing layer, which model-level defenses do not see.

Our contribution can be summarized as:

\begin{itemize}
    \item We formalize the \textit{adversarial path-selection problem}, identifying execution-path activation as the dominant lever for attacking pipeline efficiency.
    \item We present \texttt{AESOP}, the first path-aware efficiency-attack framework, combining vulnerability-guided path ranking with adaptive loss weighting, and separable from any single-model perturbation backend.
    \item We evaluate \texttt{AESOP} across five pipelines and a production-realistic deployment, achieving up to $2{,}407\times$ FLOPs inflation ($20\times$ over the strongest baseline) and showing that system-level defenses redirect the attack into $96.7\%$ silent data loss rather than neutralizing it.
\end{itemize}

\section{Background}



Deep learning pipeline systems can be formalized as directed acyclic graphs (DAGs)~\cite{hu2021scrooge}, where each node corresponds to a neural network component and edges denote the flow of data between stages. These systems are distinguished not only by their architectural composition but also by several critical operational characteristics that impact both performance and vulnerability. We highlight three such characteristics: data dependency, blocking and starvation that collectively define the efficiency landscape of deep learning pipelines. 



Pipeline systems are fundamentally defined by \textit{data dependency}, where outputs from upstream models become inputs to downstream models, enforcing a strict execution order (e.g., in AR face tracking, face detection must precede landmark detection, which must precede expression analysis)~\cite{10.1145/3240508.3240561}. This structure also leads to \textit{blocking of upstream components}: if a model runs inefficiently, its input queue can fill and stall earlier stages even if they are fast (e.g., inefficient semantic segmentation can eventually block the upstream object detector once the connecting queue is full)~\cite{10905030}. Conversely, it causes \textit{starvation of downstream components}, where slower stages deprive later stages of inputs, leaving them underutilized and increasing end-to-end latency (e.g., slow object recognition can idle action recognition and scene understanding)~\cite{9795869}. As a result, small inefficiencies in individual models can propagate and amplify across the pipeline, making system-level efficiency highly sensitive and motivating optimization approaches that account for holistic pipeline behavior rather than optimizing models in isolation~\cite{hu2021rim}.

A pipeline system exhibits dynamic behavior when its computational characteristics vary based on input properties. 
\noindent Existing works in dynamic deep learning systems introduce four different dynamic behaviors~\cite{sok-ravishka, rathnasuriya2025exploiting}: (1) Inputs can trigger different execution paths, e.g., activating more neurons or bypassing early exits; (2) The number of iterations (e.g., decoding steps) changes based on input complexity; (3) The volume of output (e.g., number of bounding boxes) varies, affecting later modules; and (4) Logic-based gates conditionally forward or drop inputs, altering the pipeline’s active path. Each category introduces unique efficiency vulnerabilities that can be targeted by adversarial inputs.

\section{Threat Model} \label{sec:mechanisms}

\begin{figure*}[t]
    \centering
    \includegraphics[width=0.9\textwidth]{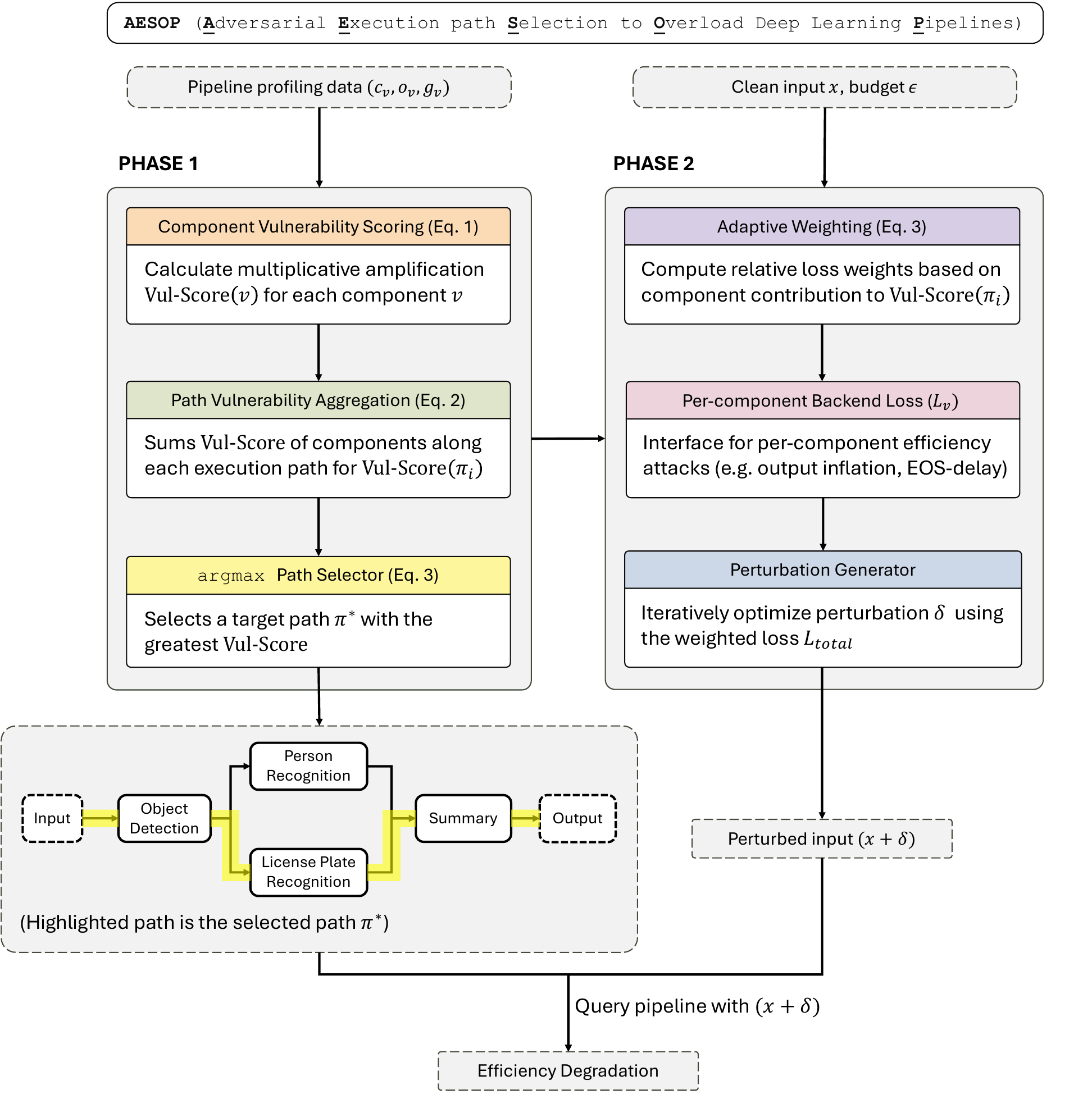}
    \caption{Approach Overview.}
    \label{fig:main_fig}
\end{figure*}

\label{sec:threat_model}


\subsection{Pipeline System Model}

We model a dynamic deep learning pipeline as a directed acyclic graph $\mathcal{S} = (\mathcal{V}, \mathcal{E})$, where each node $v \in \mathcal{E}$ is a neural component and edge edge $e \in \mathcal{E}$ carries data between components through an inter-process queue. Each component $v$ exhibits three input-dependent behaviors: a per-inference cost $c_v$, an output cardinality $o_v$ that determines downstream workload, and a gating function $g_v$ that may forward, drop, or route inputs based on predicted labels, confidence thresholds, or shape constraints. The total cost incurred at $v$ for input $x$ is $C(v, x) = c_v \cdot g_v(o_v)$. 

A end-to-end execution path $\pi \in \mathcal{P}(\mathcal{S})$ is a sequence of components activated for a given input. Because $g_v$  is input-dependent, different inputs activate different paths, and we write $\pi(x)$ for the path induced by $x$, $\pi(x) \subseteq \mathcal{P}(\mathcal{S})$. The cost of processing $x$ is therefore

\begin{equation}
    C(\mathcal{S}, x) = \sum_{v \in \pi(x)} C(v, x),
\end{equation}

which depends jointly on \textit{which} paths are activated and how expensive each component on that path becomes. When $|\mathcal{P}(\mathcal(S)| = 1$ and $|\pi(x)| = 1$, the pipeline reduces to a monolithic model, otherwise, path selection becomes an independent lever.

\subsection{Attacker's Goal.}

We consider an inference-time adversary whose objective is to degrade the \emph{efficiency} of a deployed pipeline by inflating $C(\mathcal{S}, x)$. This reflects two departures from conventional adversarial attacks: the target is efficiency rather than output integrity, and the attack surface is a multi-model pipeline rather than an individual model.

Concretely, given an input $x$ and a perturbation budget $\epsilon$, the adversary solves

\begin{equation}
    \max_{\delta:\|\delta\|_p \le \epsilon} C(\mathcal{S}, x+\delta),
\end{equation}

which decomposes into two coordinated decisions:

\begin{equation}
    \underbrace{\max_{\pi^* \in \mathcal{P}(\mathcal{S})}}_{\text{path selection}} \underbrace{\max_{\delta:\, \pi(x+\delta)=\pi^*,\, \|\delta\|_p \leq \epsilon} \sum_{v \in \pi^*} C(v,\, x+\delta)}_{\text{path-conditioned perturbation}}.
\end{equation}


Single-model efficiency attacks address only the inner maximization (path-conditioned perturbation), with the outer one (path selection) made vacuous by the assumption $|\mathcal{P}| = 1$. The pipeline setting makes both decisions consequential, and the outer one dominant. We use FLOPs as the primary efficiency metric throughout, as it is device-invariant; latency and energy are reported as secondary measures.

\subsection{Attacker's Capabilities.}
The adversary operates \textit{only at inference time} through input queries. They cannot modify model weights, access training data, alter scheduling or queueing infrastructure, or tamper with the execution environment. Perturbations are bounded under an $l_p$ constraint to preserve visual plausibility.

\subsection{Attacker's Knowledge.}We study two access levels. In the \textbf{gray-box} setting, the adversary knows the pipeline topology and can query each component to observe its inputs, outputs, and coarse runtime signals (per-component latency), but has no access to model parameters or gradients. This is sufficient to estimate $c_v$ , $o_v$, and the aggregate $g_v(o_v)$ behavior empirically, enabling Phase~1 (vulnerability-guided path ranking) path ranking, while Phase~2 (adaptive loss weighting) perturbations must be transferred from surrogate models. In the \textbf{white-box} setting, the adversary additionally has full access to model architectures, weights, and gradients, enabling direct gradient-based optimization in Phase~2.

\subsection{Out of Scope. }We focus on application-layer input attacks and exclude training-time~\cite{vassilev2024adversarial}, hardware-level, and network-based threats~\cite{mohseni2022taxonomy}, as well as attacks that modify the scheduling or queuing layer~\cite{lee2025secure}.

\section{System Design \label{sec:system-design}}

\texttt{AESOP} solves the adversarial path-selection problem in two coordinated phases:  \textit{(1) vulnerability-guided path ranking}(Section~\ref{sec:path_ranking}) profiles the pipeline, scores each execution path by the cost amplification it would incur under adversarial inputs, and selects the most vulnerable path $\pi^*$, and \textit{(2) adaptive loss weighting}(Section~\ref{sec:loss_design}) generates a perturbation against $\pi^*$ using a path-conditioned loss whose terms are weighted by each component's contribution to the path's vulnerability score. 

Figure~\ref{fig:main_fig} overviews the framework. The two phases are deliberately decoupled: Phase~1 depends only on profiling data and is shared across white-box and gray-box settings, while Phase~2 admits any single-model efficiency attack as its perturbation backend in Section~\ref{sec:separability}.

\subsection{Vulnerability-Guided Path Ranking}
\label{sec:path_ranking}
Phase~1 answers the outer maximization of the adversary's problem (Section~\ref{sec:threat_model}): given the set of execution paths $ \mathcal{P}(\mathcal{S})$, which path $\pi^*$ admits the largest cost amplification under adversarial inputs?

\subsubsection{Profiling}

For each component $v \in \mathcal{V}$, we measure the three input-dependent quantities defined in~\ref{sec:threat_model}: per-inference cost $c_v$, output cardinality $o_v$ and gating behavior $g_v$. In the white-box setting, all three are observed directly through instrumentation and code inspection. In the gray-box setting, $c_v$ is measurable via runtime profiling, while $o_v$ and $g_v$ are not individually observable but their composition $g_v(o_v)$ --- the effective workload forwarded downstream --- is recoverable from each component's input/output behavior. Phase~1 requires only the aggregate, so both settings are supported with the same procedure.

\subsubsection{Component Vulnerability Score}

For an adversarial input $x+\delta$, the cost incurred at component $v$ becomes $C^{adv}(v, x+\delta)=c^{adv} \cdot g^{adv}_v(o^{adv}_v)$, where $c^{adv}$ and $o^{adv}$ denote the adversarially induced values. We define the component-level vulnerability score as the multiplicative amplification this represents over the clean input:

\begin{equation}
    \text{Vul-Score}(v) = \frac{C^{adv}(v,\, x+\delta)}{C(v,\, x)} - 1. \label{eq:vul_score_v}
\end{equation}

A score of zero indicates a component whose cost is invariant to input perturbation (e.g., a fixed-architecture classifier with constant per-input FLOPs); large scores indicate components whose cost can be inflated substantially. 

\subsubsection{Path Vulnerability Score}

The cost incurred along a path $\pi_i \in \mathcal{P}(\mathcal{S})$  aggregates over its components, and so does the amplification:

\begin{equation}
    \text{Vul-Score}(\pi_i) = \sum_{v \in \pi_i} \text{Vul-Score}(v). \label{eq:vul_score_p}
\end{equation}

We use summation rather than maximum because pipeline cost \emph{cascades}: an inflated output at one component propagates work to every downstream component on the path simultaneously. In a traffic monitoring pipeline, a detector that emits 80 spurious bounding boxes does not just stress the detector — it forwards 80 crops to license-plate recognition, 80 queries to the knowledge-retrieval module, and 80 messages over the upload path. A maximum operator would capture only the most expensive single component and miss this multi-stage amplification, which is precisely the structural property that distinguishes pipelines from monolithic models.

The framework is robust to approximation in the score values themselves, since $\pi^* = \arg \max_{\pi^*} $  depends only on the \textit{ranking} of paths, not their absolute scores. As shown in~\ref{sec:walkthrough}, gaps between the top and runner-up paths are typically large (one to two orders of magnitude), so even coarse profiling estimates suffice to identify $\pi^*$ correctly.

\subsubsection{A Worked Example}
\label{sec:walkthrough}

We illustrate Phase~1 on a traffic monitoring pipeline (Figure~\ref{fig:main_fig}) consisting of object detection (OD), person recognition (PR), license-plate recognition (LPR), and a information summary module (SUM). The pipeline contains three execution paths, partitioned by the object detection gating decision: $\pi_{1} = \text{Input} \rightarrow OD  \rightarrow PR \rightarrow SUM \rightarrow \text{output}$, $\pi_{2} =\text{Input} \rightarrow OD \rightarrow LPR \rightarrow SUM  \rightarrow \text{output}$, and $\pi_{3} =\text{Input} \rightarrow OD \rightarrow \text{output}$. Per-component profiling yields $c_{\text{PR}} = 10.4$ GFLOPs and $c_{\text{LPR}} = 332.0$ GFLOPs a $\sim32\times$ cost asymmetry between the two downstream branches. Both branches share the same upstream detector and gating mechanism, so the workload-inflation factor available to an adversary is comparable across paths. Phase~1 therefore ranks $\pi_2$ over $\pi_1$ and selects it as $\pi^*$. This prediction is borne out empirically: targeting the car path achieves $1,254\times$ FLOPs amplification, while targeting the person path achieves only $41\times$ on the same inputs and budget --- a ratio that closely tracks the predicted per-unit cost asymmetry and confirms that path choice, not perturbation strength, dominates pipeline-level cost.

\subsection{Adaptive Loss Weighting}
\label{sec:loss_design}
Phase~2 answers the inner maximization: given the selected path $\pi^*$, generate a perturbation $\delta$ that maximizes its cost. The challenge is that $\pi^*$ contains multiple dynamic components, each with a different vulnerability profile, and naive equal-weighted combination wastes gradient signal on terms that contribute little to total amplification.

\subsubsection{Vulnerability-Weighted Objective}
We define the path-conditioned loss as a weighted sum over components on $\pi^*$, with weights set by each component's relative contribution to the path's vulnerability score:

\begin{equation}
    \mathcal{L}_{\text{total}} = \sum_{v \in \pi^*} \frac{\text{Vul-Score}(v)}{\text{Vul-Score}(\pi^*)} \cdot \mathcal{L}_v. \label{eq:weighted_loss}
\end{equation}

The weights are not free hyper-parameters: they are computed from the same Phase~1 profiling that selected $\pi^*$, so the loss adapts automatically to the topology and cost structure of the target pipeline.

\subsubsection{Per-Component Backend}
\label{sec:separability}

Each component $v \in \pi^*$ contributes a loss term $\mathcal{L}_v$ that drives perturbation gradients toward inflating that component's contribution to path cost. \texttt{AESOP} specifies only the interface: $\mathcal{L}_v$ must produce gradients that increase $C^{\text{adv}}(v, x+\delta)$. The choice of perturbation mechanism is left open, and is determined by which input-dependent behavior of $v$ admits the largest amplification and what differentiable signal is available at that node.

This decoupling is deliberate. Single-model efficiency attacks have proliferated across component types — output inflation for detectors~\cite{shapira2023phantom, chen2024overload, ma2024slowtrack}, EOS-delay for auto-regressive decoders~\cite{chen2022nmtsloth, haque2023slothspeech}, early-exit suppression for adaptive networks~\cite{hong2020panda, pan2022gradauto, pan2023gradmdm} — and each operates on a different lever with different gradient signals. Any of these can serve as a per-component backend in \texttt{AESOP} without modification to Phase~1.

The path-selection layer above these backend is what \texttt{AESOP} contributes, and is orthogonal to perturbation strength: a stronger $\mathcal{L}_v$ improves the achievable cost along $\pi^*$, but the selection of $\pi^*$ --- which our evaluation shows is the dominant factor in section~\ref{sec:whitebox} --- is supplied by a layer that single-model methods structurally do not provide.

\section{Implementation}

\begin{figure}[t]
    \begin{subfigure}[b]{0.5\textwidth}
        \centering
        \includegraphics[width=\linewidth]{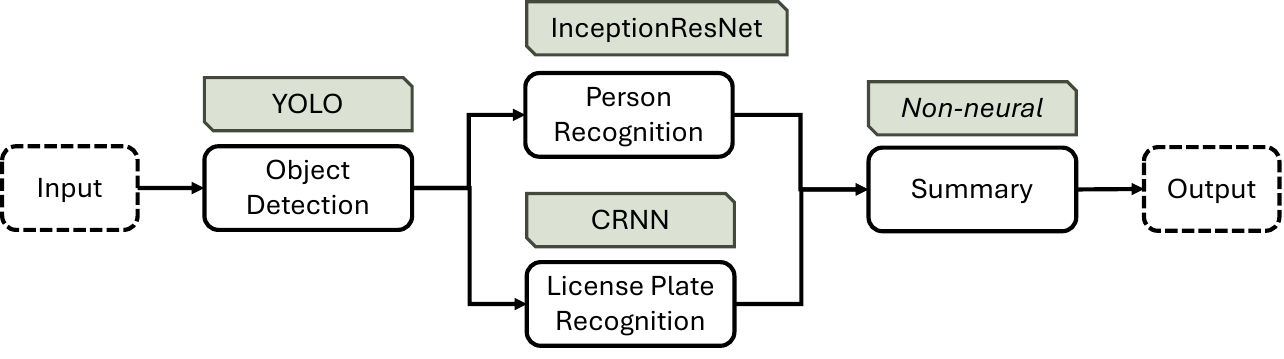}
        \caption{Baseline traffic monitoring pipeline. YOLO detects objects; \textit{person} detections go to InceptionResNet, \textit{car} detections to a CRNN-based plate reader. A summary module aggregates both branches.}

        \vspace{2em}
        \label{fig:traffic_monitoring}
    \end{subfigure}
    \begin{subfigure}[b]{0.5\textwidth}
        \centering
        \includegraphics[width=\linewidth]{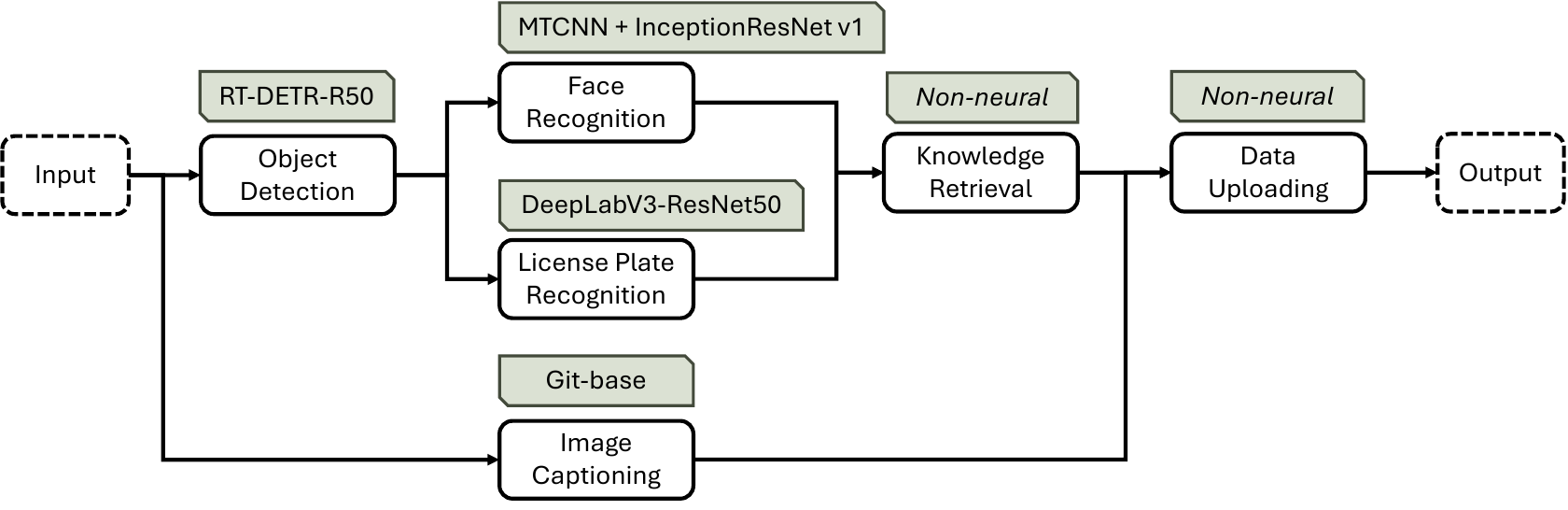}

        \caption{Production-realistic variant. RT-DETR-R50 replaces YOLO. Person and vehicle branches use MTCNN+InceptionResNet-v1 and DeepLabV3-ResNet50 respectively. A parallel GIT-base branch captions each frame, and a knowledge-retrieval module matches results against a SQLite3 database.}
        \label{fig:traffic_monitoring_var}
        \vspace{1.5em}
    \end{subfigure}

    \caption{Traffic-monitoring pipeline in two configurations. Shaded labels mark the model used by each module; \textit{Non-neural} modules contain no deep learning model and has negligible computational cost.}
\end{figure}

\subsection{Pipeline Applications}
We construct five deep learning pipeline applications and one production-realistic variant using publicly available models. The five baseline applications use untrained model weights, which lets us cover a diverse set of pipeline topologies, but limits the validity of any conclusion that depends on model accuracy. To address this, we additionally build a production-realistic variant of the traffic-monitoring pipeline using fully trained, deployment-grade models.

\textbf{Baseline traffic monitoring (Figure~\ref{fig:traffic_monitoring}).}
The pipeline captures vehicle and pedestrian information from a frame stream and writes log entries to disk or uploads them to the cloud. It integrates four components. An object detector based on YOLO produces bounding boxes with class labels; detections labeled \textit{person} are forwarded to a person-recognition module built on InceptionResNet, while detections labeled \textit{car} are sent to a license-plate-recognition module that combines a traditional (non-neural) OCR stage with a CRNN. A summary module collects the results from both branches and writes them to log files. Referred as traffic monitoring in the rest of the paper.

\textbf{Production-realistic variant (Figure~\ref{fig:traffic_monitoring_var}).}
The variant preserves the same routing structure but replaces every neural component with a fully trained, deployment-grade counterpart. The frame stream is processed by an RT-DETR detector chosen for its higher throughput. \textit{Person} detections are forwarded to MTCNN for face localization followed by Inception-ResNet-v1 for face recognition; \textit{car} detections are forwarded to DeepLabV3 for license-plate segmentation and recognition. Both branches feed a knowledge-retrieval module that uses a GPT-2 tokenizer to match recognized faces and plates against entries in a SQLite3 database. In parallel with the object detector, a GIT-base captioning model generates a natural-language description of each frame. At the pipeline endpoint, a data-uploading component transmits all results to the cloud at negligible computational cost. We refer to this pipeline as \textit{traffic monitoring} in the rest of the paper. We refer to this pipeline as \textit{traffic variant} in the rest of the paper.

The remaining four pipeline applications (wildlife conservation, smart-home alerting, expressway surveillance, and Amber Alert system) follow the same construction style and are described in Figure~\ref{fig:five_applications}.

\subsection{Datasets}

\textbf{MS COCO.} Following prior work~\cite{chen2024overload, ma2024slowtrack, shumailov2021sponge}, we use MS COCO as the primary evaluation dataset for four of the five pipelines.

\textbf{Animal Faces.} The wildlife conservation pipeline is evaluated on the Kaggle Animal Faces dataset, which contains approximately 15{,}000 samples across three classes (cat, dog, and wild).

\subsection{Metrics}
\label{sec:metrics}
We report per-module and end-to-end \textbf{FLOPs} (average floating-point operations over the evaluation set) and \textbf{latency} (average processing time over the evaluation set). We additionally report \textbf{workload}, defined as the number of requests forwarded to a component per system input, which captures the cascading inflation that drives system-level cost. We use FLOPs as the primary metric throughout, as it is device-invariant; latency and workload are reported as secondary measures and are sensitive to hardware and scheduling state.

\subsection{System Environment}

\noindent \textbf{Hardware.} Adversarial examples are generated on a server with 4$\times$ RTX 3090 (24GB) and an Intel Xeon E5-2686 v4 , running CUDA 12.2 with driver 535.183.01. Pipeline experiments use the same machine with a single GPU to reflect realistic single-device deployment, except the Qwen2-70B-Instruct model is hosted on a 4$\times$A100 (80GB) machine. 

\noindent \textbf{Hyperparameters.} Unless otherwise noted, we run 400 attack iterations with an $\ell_\infty$ budget of 4\% and fixed loss coefficients; and fixed loss coefficients. Results are insensitive to these settings within reasonable ranges.

\section{Evaluation}

We organize the evaluation around four research questions, each isolating a specific claim from Section~\ref{sec:system-design}.

\textbf{RQ1 (White-box effectiveness; Section~\ref{sec:whitebox}).} Given identical inputs, perturbation budgets, and gradient access, does path-aware targeting amplify pipeline cost beyond what perturbation-only methods can achieve? RQ1 tests the core thesis: path selection is a lever separate from perturbation strength.

\textbf{RQ2 (Gray-box transfer; Section~\ref{sec:graybox}).} Does the path-selection mechanism survive without gradient access to the target? Phase~1 ranking is reproduced from public model specifications; Phase~2 perturbations are generated against surrogates and transferred. RQ2 distinguishes a structural property of pipelines from an artifact of white-box capability.

\textbf{RQ3 (Component-wise contribution; Section~\ref{sec:ablation}).} How much of the amplification comes from Phase~1 (path ranking) versus Phase~2 (perturbation generation)? Reduced variants disable path selection while preserving the perturbation backend, isolating the contribution of the algorithmic primitive that single-model attacks lack.

\textbf{RQ4 (Production-realistic deployment; Section~\ref{sec:production}).} Does the path-selection vulnerability persist under deployment configurations and defenses approximating production systems? We evaluate bounded buffering, batched scheduling, confidence thresholding, input preprocessing, and ML-based input filtering. RQ4 tests whether the attack is an artifact of idealized evaluation or survives system-level mitigation.

The four questions are sequenced from idealized to deployed: RQ1 establishes the maximum advantage of path selection, RQ2 removes target-gradient access, RQ3 attributes the advantage to specific framework components, and RQ4 subjects the surviving attack to production deployment conditions.

\subsection{White-box Effectiveness}
\label{sec:whitebox}

\begin{table*}[htbp]
  \centering
  \caption{Performance of Attacks on Different Pipelines. FLOPs$\times$ and Latency$\times$ denote multiplicative increases in end-to-end computational cost and runtime relative to benign inputs. The unit of latency is second.}
  \scriptsize \renewcommand{\arraystretch}{1}
  \tiny
  \resizebox{0.8\textwidth}{!}{
    \begin{tabular}{clrrrrrr}
    \toprule
    \textbf{Application} & \textbf{Method}  & \textbf{Latency} & \textbf{Latency$\times$} & \textbf{GFLOPs} & \textbf{FLOPs$\times$} \\
    \midrule
    \multirow{\textbf{Traffic Monitoring}} & Benign   & 0.045 & 1.00 & 10.31 & 1.00 \\
     & Overload  & 2.492 & 55.38 & 1063.23 & 103.13 \\
     & Phantom  & 0.536 & 11.91 & 130.33 & 12.64 \\
     & SlowTrack   & 2.127 & 47.27 & 632.43 & 61.34 \\
     &  \textbf{\tool} & \textbf{13.654} & \textbf{303.42} & \textbf{3389.76} & \textbf{328.78} \\
    \midrule
    \multirow{\textbf{Wildlife Conservation}} & Benign  & 0.021 & 1.00 & 2.59 & 1.00 \\
     & Overload  & 0.209 & 9.95 & 56.83 & 21.94 \\
     & Phantom & 0.050 & 2.38 & 2.48 & 0.96 \\
     & SlowTrack  & 0.188 & 8.95 & 24.21 & 9.35 \\
     &  \textbf{\tool}   & \textbf{8.803} & \textbf{419.19} & \textbf{5251.35} & \textbf{2027.55} \\
    \midrule
    \multirow{\textbf{Smart Home}} & Benign & 5.740 & 1.00 & 3.25 & 1.00 \\
     & Overload  & 6.430 & 1.12 & 24.23 & 7.46 \\
     & Phantom  & 6.785 & 1.18 & 2.37 & 0.73 \\
     & SlowTrack  & 6.180 & 1.08 & 23.58 & 7.26 \\
     &  \textbf{\tool} & \textbf{11.672} & \textbf{2.03} & \textbf{2828.46} & \textbf{870.30} \\
    \midrule
    \multirow{\textbf{Expressway Surveillance}} & Benign  & 0.022 & 1.00 & 5.80 & 1.00 \\
     & Overload  & 0.316 & 14.36 & 635.23 & 109.52 \\
     & Phantom  & 0.047 & 2.14 & 8.00 & 1.38 \\
     & SlowTrack  & 0.355 & 16.14 & 678.13 & 116.92 \\
     & \textbf{\tool} & \textbf{5.796} & \textbf{263.45} & \textbf{13963.75} & \textbf{2407.54} \\
    \midrule
    \multirow{\textbf{Amber Alert}} & Benign  & 5.044 & 1.00 & 122.87 & 1.00 \\
     & Overload & 6.386 & 1.27 & 13787.53 & 112.21 \\
     & Phantom  & 3.736 & 0.74 & 2795.27 & 22.75 \\
     & SlowTrack  & 5.782 & 1.15 & 11789.03 & 95.95 \\
     & \textbf{\tool} & \textbf{24.743} & \textbf{4.91} & \textbf{76469.74} & \textbf{622.36} \\
    \bottomrule
    \end{tabular}%
    }
  \label{tab:performance_of_attacks_on_different_pipelines}%
\end{table*}%

\begin{table}[htbp]
  \centering
  \caption{Gray-box Transferability. Workload denotes the number of detections processed by all direct downstream modules of the object detector. FLOPs$\times$ and Latency$\times$ denote multiplicative increases in end-to-end computational cost and runtime relative to benign inputs. The unit of latency is second.}
  \huge
  \resizebox{0.48\textwidth}{!}{
    \begin{tabular}{clrrrrr}
    \toprule
    \textbf{Model} & \textbf{Application} & \textbf{Workload} & \textbf{Latency} & \textbf{Latency $\times$} & \textbf{FLOPs} & \textbf{FLOPs $\times$} \\
    \midrule
    \multirow{\textbf{YOLOv5l}} & Traffic & 36.70  & 0.52  & 11.68  & 270.35  & 4.15  \\
          & Wildlife & 59.02  & 0.40  & 19.25  & 297.17  & 5.41  \\
          & Smart Home  & 27.95  & 7.40  & 1.29  & 214.07  & 3.81  \\
          & Expressway & 64.86  & 0.37  & 17.03  & 977.07  & 16.27  \\
          & Amber Alert & 36.65  & 3.79  & 0.75  & 1846.20  & 12.86  \\
    \midrule
    \multirow{\textbf{YOLOv8n}} & Traffic & 12.97  & 0.21  & 4.62  & 45.28  & 3.60  \\
          & Wildlife & 10.40  & 0.07  & 3.48  & 47.11  & 9.85  \\
          & Smart Home  & 4.09  & 4.89  & 0.85  & 21.18  & 5.18  \\
          & Expressway & 33.66  & 0.21  & 9.46  & 483.17  & 58.25  \\
          & Amber Alert & 12.96  & 3.85  & 0.76  & 925.50  & 7.30  \\
    \bottomrule
    \end{tabular}%
    }
  \label{tab:graybox_transferability}%
\end{table}%

\begin{figure}[t]
    \centering
    \includegraphics[width=0.5\textwidth]{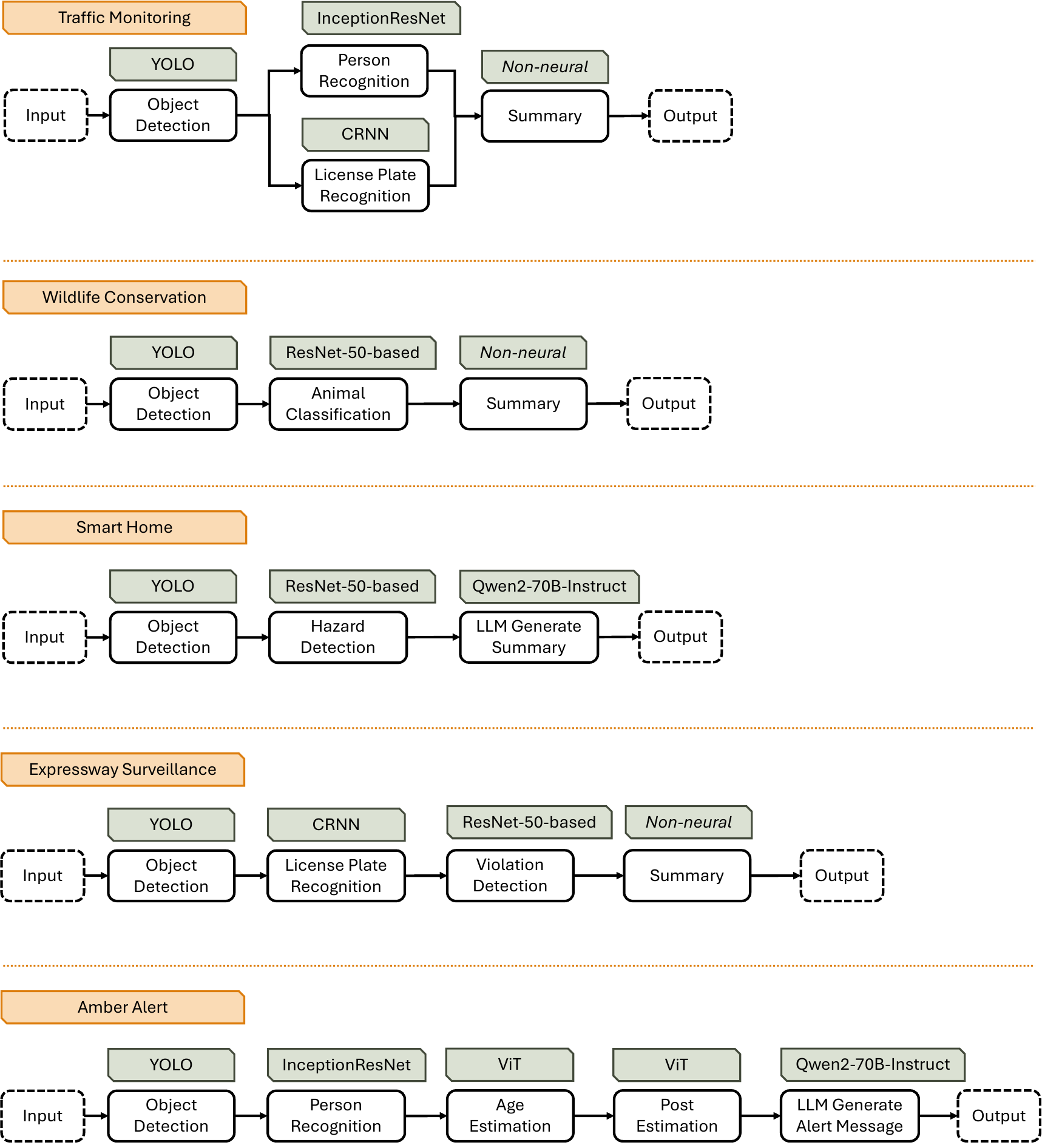}
    \caption{Pipeline applications used in evaluation. Implementation details are stated in Appendix~\ref{sec:pipeline_applications}.}
    \label{fig:five_applications}
\end{figure}

\textbf{Setup.}
This experiment isolates path-aware targeting from perturbation strength. All methods see identical inputs, an identical $l_\infty$ budget of 4\%, identical 400-iteration optimization, and identical surrogate access. Only the choice of target and the loss shaping differ. The adversary has full white-box access in both Phase~1 (path ranking) and Phase~2 (perturbation generation). We compare \tool\ against three single-model efficiency attacks: Overload~\cite{chen2024overload}, Phantom Sponges~\cite{shapira2023phantom}, and SlowTrack~\cite{ma2024slowtrack}. Each baseline attacks the object detector at the head of every pipeline, the standard single-model convention~\cite{chen2024overload}. Baselines receive no pipeline-level information; they were not designed to use it.

\textbf{Pipelines with LLM components dampen latency multipliers, not FLOPs.}
Two of the evaluated pipelines, Smart Home and Amber Alert (Figure~\ref{fig:five_applications}), terminate in a large language model that dominates clean-input runtime. The LLM consumes a structured JSON summary of upstream outputs, which bounds prompt size regardless of how much workload upstream modules produce. Amplification at the detection head therefore adds wall-clock time only in proportion to the upstream share of total runtime, which is small. On Smart Home, \tool\ produces 870$\times$ FLOPs amplification with only 2.03$\times$ latency; on Amber Alert, 622$\times$ FLOPs and 4.91$\times$ latency. This is a property of the pipeline, not a weakness of the attack. The operational implication is sharper than the latency numbers alone suggest: the attack remains invisible at the SLA layer (a few seconds slower) while the compute bill grows two to three orders of magnitude. The same effect appears under transfer (Section~\ref{sec:graybox}) and matches prior observations on autoregressive bottlenecks~\cite{chen2022nmtsloth}.

\textbf{\tool\ amplifies pipeline workload by 622--2{,}407$\times$ while no baseline exceeds 117$\times$.}
Across the five pipelines, \tool\ achieves 622--2{,}407$\times$ FLOPs amplification, while the strongest baseline on any pipeline reaches only 117$\times$ (Table~\ref{tab:performance_of_attacks_on_different_pipelines}). The largest absolute amplification, 2{,}407$\times$ FLOPs and 263$\times$ latency on Expressway Surveillance, exceeds the strongest baseline on that pipeline (SlowTrack at 117$\times$) by 21$\times$. The relative gap to the strongest baseline ranges from 3.2$\times$ on Traffic Monitoring (paths are nearly symmetric) to 117$\times$ on Smart Home (paths are highly asymmetric). This gap is not a difference in perturbation strength: budgets, iterations, and surrogate access are identical. \tool\ directs the perturbation against the most expensive downstream path; baselines inflate detector activity without regard to where it is routed.

\textbf{The size of the gap tracks how asymmetric the pipeline's paths are.}
Wildlife Conservation is the clearest case. Only detections labeled \textit{elephant} reach the fine-grained classifier; every other class exits at the gate. The elephant path is the most expensive and the least activated under clean inputs, so a path-blind attack inflates whichever class the detector already responds to and fails to push workload into the costly branch. Smart Home repeats the pattern: the LLM-bound hazard path dominates an early-exit alternative, and a path-blind attack routes most workload into the early-exit branch where it is structurally bounded. Traffic Monitoring sits at the other end of the spectrum. Face recognition and license plate recognition have comparable per-unit cost in the simulated configuration, so target choice matters less and the gap shrinks to 3.2$\times$. The floor of the gap is roughly 3$\times$, the ceiling two orders of magnitude.

\textbf{Path-blind attacks can backfire on asymmetric pipelines.}
Phantom Sponges produces 0.96$\times$ FLOPs on Wildlife Conservation and 0.73$\times$ on Smart Home, both below the clean baseline. Phantom is designed to flood NMS survivors, which inflates cost only when NMS-bound post-processing is the bottleneck. In Wildlife the bottleneck is the fine-grained classifier downstream of detection; in Smart Home it is the LLM. Phantom's perturbations succeed at the detector head but cannot propagate cost into the modules that dominate runtime. The result is that a stronger detector-level attack delivers no additional pipeline-level cost, sometimes less. This is the failure mode that motivates path-aware attack design: the question of \emph{which} module to flood is logically prior to the question of \emph{how} to flood it.

\begin{tcolorbox}[breakable]
\textbf{Answer to RQ1.}
Path-aware targeting amplifies pipeline cost by 622--2{,}407$\times$ FLOPs across the five pipelines, beyond what any perturbation-only baseline achieves on the same inputs. Even where baselines reach nontrivial amplification (Overload at 103$\times$ on Traffic Monitoring, SlowTrack at 117$\times$ on Expressway Surveillance), \tool\ outperforms them by 3.2--117$\times$ under identical budgets and surrogate access. Path selection compounds with the perturbation backend rather than competing with it: any improvement to single-model efficiency attacks would inflate the absolute numbers reported here without closing the gap. What baselines are missing is not perturbation strength but an algorithmic primitive: the choice of which path to attack.
\end{tcolorbox}


\subsection{Gray-box Transfer Effectiveness}
\label{sec:graybox}

\textbf{Setup.}
This experiment tests whether \tool\ retains its path-aware advantage when the adversary loses gradient access to the target. We split the framework at the access boundary defined in Section~\ref{sec:threat_model}. Phase~1 (path ranking) uses public model specifications, available to any gray-box adversary. Phase~2 (perturbation) is generated against surrogate models and transferred. The surrogate is YOLOv5m; targets are YOLOv5l (same family, larger capacity) and YOLOv8n (different family, smaller capacity). Perturbation budget, optimization iterations, and pipeline configurations match Section~\ref{sec:whitebox}. Only the gradient source changes.

\textbf{Phase~1 ranking is reproducible from public model specifications.}
Phase~1 ranking depends on the relative ordering of paths, not exact cost values. In every evaluated pipeline, the most expensive path exceeds the cheapest by more than an order of magnitude. In the production-realistic Traffic variant, for example, DeepLabV3-ResNet50 with $\sim$332~GFLOPs~\cite{teng_deeplabv3_2019} on the LPR path is roughly 32$\times$ more expensive than MTCNN + InceptionResnetV1 ($\sim$10~GFLOPs) on the FR path~\cite{TERMRITTHIKUN2019579}.
Both numbers are recoverable from FLOPs counts reported alongside each architecture in standard model repositories. No probing of the deployed system is required. Phase~1 is therefore a public-information lookup. What public information cannot supply, and what the gray-box experiments below evaluate, is the per-input perturbation.

\textbf{Path-aware perturbations transfer with 3.6--58.3$\times$ FLOPs amplification.}
\tool\ achieves 3.6--58.3$\times$ FLOPs and up to 19.3$\times$ latency amplification across the five pipelines and two target architectures (Table~\ref{tab:graybox_transferability}). Workload counts of 28--65 detections per frame on YOLOv5l confirm the mechanism: surrogate-crafted perturbations still steer the target detector into producing the path-vulnerable class. The magnitudes are smaller than the white-box numbers, as expected, since transfer is a strict capability reduction. The path-selection mechanism survives in every pipeline.

\textbf{Same-family transfer is stronger on workload and latency.}
Workload transfer is larger on YOLOv5l (28--65 detections per frame) than on YOLOv8n (4--34). Perturbations crafted on YOLOv5 surrogates align more tightly with same-family detection heads, as expected. Latency follows the same pattern: 1.3--19.3$\times$ on YOLOv5l, 0.8--9.5$\times$ on YOLOv8n. Amber Alert is a partial exception (latency near unity for both targets), attributable to the LLM dampening described in Section~\ref{sec:whitebox}.

\textbf{A smaller, faster target is not a more robust target.}
The cross-family target produces \emph{higher} relative FLOPs amplification than the same-family target on three of the five pipelines, despite weaker absolute workload transfer. On Expressway Surveillance, YOLOv8n shows 58.3$\times$ FLOPs versus YOLOv5l's 16.3$\times$; on Wildlife Conservation, 9.9$\times$ versus 5.4$\times$; on Smart Home, 5.2$\times$ versus 3.8$\times$. The inversion follows from arithmetic: YOLOv8n is the cheaper detector, so the same absolute induced workload divides into a smaller denominator. The deployment implication is that a faster, cheaper detector is not a more robust target. It is one whose pre-attack baseline begins lower, leaving more headroom for an attacker to inflate.

\begin{tcolorbox}[breakable]
\textbf{Answer to RQ2.}
The path-selection mechanism survives the loss of target-gradient access in every pipeline tested. Phase~1 reduces to a public-information lookup: relative path costs are recoverable from architecture-level FLOPs counts in standard model repositories, with no probing of the deployed system. Phase~2 transfers from a YOLOv5m surrogate to both same-family (YOLOv5l) and cross-family (YOLOv8n) targets at 3.6--58.3$\times$ FLOPs amplification. White-box access sharpens path-aware attacks; it is not a precondition. An adversary with public model specifications and a same-family surrogate can execute the full attack pipeline against an unseen target.
\end{tcolorbox}


\subsection{Component-wise Contribution Analysis}
\label{sec:ablation}

\begin{table}[h]
  \centering
  \caption{Ablation Study. FLOPs$\times$ and Latency$\times$ denote multiplicative increases in end-to-end computational cost and runtime relative to benign inputs. The unit of latency is second.}
  \huge
  \resizebox{0.48\textwidth}{!}{
    \begin{tabular}{lrrrrrr}
    \toprule
    
    \multicolumn{7}{l}{\textbf{Traffic Monitoring}} \\

          & \multicolumn{1}{l}{\textit{Person}} & \multicolumn{1}{l}{\textit{Car}} & \multicolumn{1}{l}{\textbf{Latency}} & \multicolumn{1}{l}{\textbf{Latency $\times$}} & \multicolumn{1}{l}{\textbf{FLOPs}} & \multicolumn{1}{l}{\textbf{FLOPs $\times$}} \\
    \midrule
    Benign & 1.69  & 0.25  & 0.04  & 1.00  & 10.10  & 1.00  \\
    \approach -system & 935.55  & 119.61  & 11.97  & 267.26  & 4158.20  & 411.56  \\
    \approach -system(\textit{Person}) & 1075.55  & 0.02  & \textbf{13.65}  & \textbf{304.79}  & 3389.76  & 335.51   \\
    \textbf{\tool} & 0.29  & 981.54  & 3.91  & 87.30  & \textbf{9929.04}  & \textbf{982.74}   \\
    \midrule

    \addlinespace[30pt]
    \midrule
    
    \multicolumn{7}{l}{\textbf{Wildlife Surveillance}} \\
          & \multicolumn{1}{l}{\textit{Giraffe}} & \multicolumn{1}{l}{\textit{Elephant}} & \multicolumn{1}{l}{\textbf{Latency}} & \multicolumn{1}{l}{\textbf{Latency $\times$}} & \multicolumn{1}{l}{\textbf{FLOPs}} & \multicolumn{1}{l}{\textbf{FLOPs $\times$}} \\
    \midrule
    Benign & 0.03  & 0.06  & 0.02  & 1.00  & 2.59  & 1.00  \\
    \approach -system & 134.25  & 1070.18  & 7.49  & 357.22  & 4954.25  & 1914.54  \\
    \approach -system(\textit{Giraffe}) & 866.92  & 0.01  & 5.58  & 266.37  & 3566.65  & 1378.31  \\
    \textbf{\tool} & 0.00  & 1276.69  & \textbf{8.80}  & \textbf{419.93}  & \textbf{5251.35} & \textbf{2029.35}  \\
    \bottomrule
    \end{tabular}
    }
  \label{tab:ablation_study_on_traffic_and_wildlife}
\end{table}

\textbf{Setup.}
This experiment isolates Phase~1 path selection by comparing \tool\ against two reduced variants. \tool-system removes path selection and optimizes all class-related terms uniformly, giving the framework no signal about which path is more expensive. \tool-system($\cdot$) forces the framework to attack the \emph{alternative} class, the one path selection would have rejected. Both variants retain the Phase~2 perturbation backend; only the directional control is disabled. We evaluate both on Traffic Monitoring and Wildlife Surveillance with all other settings identical to Section~\ref{sec:whitebox}. Results appear in Table~\ref{tab:ablation_study_on_traffic_and_wildlife}.

\textbf{Path selection contributes most where the detector's default class opposes the path-vulnerable class.}
The contribution of path selection depends on whether the framework's optimal target aligns with the detector's natural class bias. On Wildlife the two coincide: the framework picks Elephant, also the class the detector responds to most strongly under clean inputs, and removing path selection costs only 6\% of the amplification (1{,}915$\times$ for no-selection vs.\ 2{,}029$\times$ for full \tool). On Traffic the two oppose: the framework picks Car, the LPR-bound path at $\sim$332~GFLOPs/unit~\cite{TERMRITTHIKUN2019579}, roughly 33$\times$ more expensive than the FR-bound Person path at $\sim$10~GFLOPs/unit~\cite{teng_deeplabv3_2019}. But detectors are typically more confident on Person than Car under clean traffic imagery, so removing path selection costs 58\% of the amplification (412$\times$ vs.\ 983$\times$). Forcing the framework onto the rejected target drops amplification further, to 1{,}378$\times$ on Wildlife and 336$\times$ on Traffic, below even no-selection. The 1.5--2.9$\times$ gap between full \tool\ and the wrong-target variant controls for natural bias and isolates the directional choice. Path selection matters most where it is hardest to substitute for, where the detector's default class is misaligned with the path-vulnerable class.

\textbf{On Traffic, FLOPs and latency disagree about which target is worst.}
\tool\ achieves 87$\times$ latency and 983$\times$ FLOPs on Traffic by attacking Car; the wrong-target variant achieves 305$\times$ latency and 336$\times$ FLOPs by attacking Person. The Person path generates many low-cost FR calls ($\sim$10~GFLOPs each) that the runtime dispatches largely serially because of per-call CPU overhead. The Car path generates fewer high-cost LPR calls ($\sim$332~GFLOPs each) that GPUs pipeline efficiently. Person therefore inflates wall-clock time despite consuming less total compute. \tool\ optimizes the device-invariant metric defined as primary in Section~\ref{sec:threat_model}, FLOPs, and selects Car on this objective. On Wildlife the two metrics agree and the divergence does not arise. Section~\ref{sec:production} returns to this point under production deployment, where bounded buffers and batched scheduling shift the bottleneck and make total compute a more direct predictor of system-level harm than per-image latency.

\begin{tcolorbox}[breakable]
\textbf{Answer to RQ3.}
The decomposition is asymmetric and depends on whether the detector's natural class bias aligns with the path-vulnerable class. On Wildlife, where the two coincide (Elephant is both path-vulnerable and the detector's most confident class under clean inputs), Phase~2 alone recovers 94\% of full \tool's amplification (1{,}915$\times$ vs.\ 2{,}029$\times$); Phase~1 contributes the remaining 6\%. On Traffic, where the two oppose (Car is path-vulnerable but Person is the natural bias), Phase~2 alone recovers only 42\% (412$\times$ vs.\ 983$\times$); Phase~1 contributes the remaining 58\%. The 1.5--2.9$\times$ gap between full \tool\ and the wrong-target variant controls for natural bias and isolates the directional choice. Path selection is therefore not a fixed-percentage boost; its value scales with how badly the perturbation backend would otherwise misroute its output. It is the algorithmic primitive that single-model attacks structurally cannot supply.
\end{tcolorbox}

\subsection{Effectiveness Under Production-Realistic Deployment and Defenses}
\label{sec:production}

\begin{table*}[h]
\centering
\caption{Applying gaussian noise, spatial smoothing, and SVM input filtering on clean inputs to test defense overhead. Total number of inputs is 100. The unit of throughput is input per second; the unit of average end-to-end latency (AvgE2E), p50, p95, p99 is second; LPR\# denotes the amount of workload processed by the license-plate-recognition module; Drops denotes the number of workload dropped by bounded buffering.}
\label{tab:clean-side-effect}
\begin{tabular}{lrrrrrrrrr}
\toprule
\textbf{Defense} & \textbf{Wall Time(s)} & \textbf{Throughput} & \textbf{AvgE2E} & \textbf{p50} & \textbf{p95} & \textbf{p99} & \textbf{LPR\#} & \textbf{Drops} & \textbf{Total FLOPs} \\ \midrule
none   & 151.1 & 0.662 & 37.38 & 22.65 & 113.34 & 116.09 & 64 & 0 & 33.0T \\
gauss  & 154.3 & 0.648 & 38.72 & 23.44 & 115.49 & 118.44 & 63 & 0 & 32.6T \\
smooth & 158.0 & 0.633 & 38.20 & 21.13 & 112.44 & 114.71 & 70 & 0 & 35.7T \\
svm    & 150.7 & 0.664 & 36.84 & 21.39 & 111.50 & 114.07 & 64 & 0 & 33.0T \\ \bottomrule
\vspace{1em}
\end{tabular}
\end{table*}

\begin{table*}[h]
\centering
\caption{Input defenses on 10 perturbed inputs. The unit of throughput is input per second; the unit of average end-to-end latency (AvgE2E), p50, p95, p99 is second; LPR\# denotes the amount of workload processed by the license-plate-recognition module; Filtered denotes the number of workload filtred by SVM.}
\label{tab:attacked10}
\begin{tabular}{lrrrrrrrrr}
\toprule
\textbf{Defense} & \textbf{Wall Time(s)} & \textbf{Throughput} & \textbf{AvgE2E} & \textbf{p50} & \textbf{p95} & \textbf{p99} & \textbf{LPR\#} & \textbf{Drops} & \textbf{Total FLOPs} \\ \midrule
gauss  & 213.4 & 0.047 & 108.82 & 106.51 & 194.13 & 194.15 & 850 & 0 & 18.2T \\
smooth & 512.7 & 0.020 & 299.48 & 260.82 & 465.95 & 477.70 & 1812 & 0 & 6.6T \\
svm    & 529.2 & 0.019 & 406.42 & 406.42 & 504.76 & 513.51 & 1895 & 8 & 25.4T \\ \bottomrule
\vspace{1em}
\end{tabular}
\end{table*}
\begin{table*}[ht]
\centering
\caption{Randomly mix clean inputs with perturbed inputs. Total number of inputs is 100, covers ratio: clean:perturbed=9:1, 9.5:0.9 and 9.9:0.1. Condition denotes the defense technique applied, respectively: no defense, gaussian noise, spatial smoothing, SVM input filtering, and the combination of batch size 16 and confidence threshold 0.5. The unit of throughput is input per second; the unit of average end-to-end latency (AvgE2E), p50, p95, p99 is second; LPR\# denotes the amount of workload processed by the license-plate-recognition module; Filtered denotes the number of workload filtred by SVM. Mean $\pm$ std over 5 random seeds.}
\label{tab:mix-study}
\begin{tabular}{llrrrrrrrrr}
\toprule
\textbf{Ratio} & \textbf{Config} & \textbf{Wall Time(s)} & \textbf{Throughput} & \textbf{AvgE2E} & \textbf{p99} & \textbf{LPR\#} & \textbf{Filtered} & \textbf{Total FLOPs} \\ \midrule
9:1 & none  & 2725.7 $\pm$ 395.8 & 0.037 $\pm$ 0.007 & 137.44 $\pm$ 26.61 &  888.59 $\pm$ 149.24 & 9325 $\pm$ 85 & 0.0 $\pm$ 0.0 & 294.2T \\
9:1 & gauss  & 640.2 $\pm$ 118.6 & 0.160 $\pm$ 0.029 & 99.23 $\pm$ 8.84 &  521.54 $\pm$ 73.80 & 2058 $\pm$ 379 & 0.0 $\pm$ 0.0 & 86.0T \\
9:1 & smooth  & 547.5 $\pm$ 28.1 & 0.183 $\pm$ 0.010 & 104.19 $\pm$ 13.41 & 489.38 $\pm$ 45.72 & 1874 $\pm$ 6 & 0.0 $\pm$ 0.0 & 79.3T \\
9:1 & svm  & 573.0 $\pm$ 11.4 & 0.175 $\pm$ 0.003 & 88.32 $\pm$ 14.68 &  532.76 $\pm$ 11.99 & 1950 $\pm$ 6 & 8.0 $\pm$ 0.0 & 87.4T \\
9:1 & b16 + conf.5  & 907.5 $\pm$ 64.8 & 0.111 $\pm$ 0.008 & 49.90 $\pm$ 6.64 &  371.88 $\pm$ 13.55 & 6610 $\pm$ 692 & 0.0 $\pm$ 0.0 & 235.9T \\ \midrule 
9.5:0.5 & none  & 1421.8 $\pm$ 77.5 & 0.071 $\pm$ 0.004 & 97.78 $\pm$ 13.30 &  881.92 $\pm$ 51.66 & 4647 $\pm$ 57 & 0.0 $\pm$ 0.0 & 95.4T \\
9.5:0.5 & gauss  & 307.4 $\pm$ 67.9 & 0.337 $\pm$ 0.069 & 65.75 $\pm$ 4.59 &  246.30 $\pm$ 30.76 & 960 $\pm$ 284 & 0.0 $\pm$ 0.0 & 49.2T \\
9.5:0.5 & smooth  & 287.5 $\pm$ 46.5 & 0.356 $\pm$ 0.066 & 64.01 $\pm$ 14.23 &  233.86 $\pm$ 68.76 & 843 $\pm$ 228 & 0.0 $\pm$ 0.0 & 66.4T \\
9.5:0.5 & svm  & 361.1 $\pm$ 108.0 & 0.292 $\pm$ 0.062 & 61.28 $\pm$ 9.10 &  329.92 $\pm$ 104.18 & 1201 $\pm$ 415 & 3.8 $\pm$ 0.4 & 78.7T \\
9.5:0.5 & b16 + conf.5  & 471.6 $\pm$ 69.1 & 0.216 $\pm$ 0.030 & 31.06 $\pm$ 6.56 &  323.33 $\pm$ 40.01 & 3352 $\pm$ 522 & 0.0 $\pm$ 0.0 & 75.6T \\ \midrule 
9.9:0.1 & none  & 313.5 $\pm$ 3.6 & 0.319 $\pm$ 0.004 & 64.87 $\pm$ 7.94 &  283.66 $\pm$ 5.66 & 999 $\pm$ 12 & 0.0 $\pm$ 0.0 & 58.7T \\
9.9:0.1 & gauss  & 174.2 $\pm$ 19.8 & 0.579 $\pm$ 0.058 & 46.86 $\pm$ 5.59 & 140.29 $\pm$ 19.01 & 242 $\pm$ 160 & 0.0 $\pm$ 0.0 & 46.0T \\
9.9:0.1 & smooth  & 194.7 $\pm$ 30.0 & 0.523 $\pm$ 0.080 & 54.62 $\pm$ 14.24 &  157.75 $\pm$ 34.31 & 364 $\pm$ 212 & 0.0 $\pm$ 0.0 & 45.0T \\
9.9:0.1 & svm  & 221.4 $\pm$ 98.6 & 0.523 $\pm$ 0.200 & 51.20 $\pm$ 17.00 &  184.70 $\pm$ 99.01 & 450 $\pm$ 530 & 0.6 $\pm$ 0.5 & 55.6T \\
9.9:0.1 & b16 + conf.5  & 146.9 $\pm$ 13.7 & 0.686 $\pm$ 0.072 & 17.07 $\pm$ 5.11 &  120.90 $\pm$ 15.39 & 806 $\pm$ 94 & 0.0 $\pm$ 0.0 & 55.4T \\ \bottomrule
\end{tabular}
\end{table*}

\begin{table*}[ht]
\centering
\caption{Randomly mix clean inputs with perturbed inputs. Total number of inputs is 100, covers ratio: clean:perturbed=9:1, 9.5:0.9 and 9.9:0.1. The unit of throughput is input per second; the unit of average end-to-end latency (AvgE2E), p50, p95, p99 is second; LPR\# denotes the amount of workload processed by the license-plate-recognition module; Filtered denotes the number of workload filtred by SVM.}
\label{tab:svm-per-seed}
\begin{tabular}{lrrrrrrrrrr}
\toprule
\textbf{Ratio} & \textbf{Seed} & \textbf{Wall Time(s)} & \textbf{Throughput} & \textbf{AvgE2E} & \textbf{p50} & \textbf{p95} & \textbf{p99} & \textbf{LPR\#} & \textbf{Filtered} & \textbf{Total FLOPs} \\ \midrule
9.9:0.1 & 0 & 148.5 & 0.674 & 36.31 & 21.52 & 109.73 & 112.18 & 64 & 1 & 32.8T \\
9.9:0.1 & 1 & 148.3 & 0.674 & 38.89 & 6.25 & 106.25 & 112.32 & 64 & 1 & 32.5T \\
9.9:0.1 & 2 & 330.1 & 0.303 & 73.87 & 59.71 & 290.10 & 295.04 & 1031 & 0 & 95.2T \\
9.9:0.1 & 3 & 328.6 & 0.304 & 64.83 & 6.51 & 288.64 & 291.26 & 1031 & 0 & 84.3T \\
9.9:0.1 & 4 & 151.4 & 0.660 & 42.12 & 38.66 & 112.06 & 112.69 & 61 & 1 & 33.4T \\ \midrule
9.5:0.5 & 0 & 315.1 & 0.317 & 66.92 & 21.08 & 285.88 & 289.83 & 1026 & 4 & 32.2T \\
9.5:0.5 & 1 & 311.1 & 0.321 & 55.22 & 20.42 & 138.66 & 280.19 & 1021 & 4 & 41.1T \\
9.5:0.5 & 2 & 312.0 & 0.320 & 70.11 & 49.74 & 281.47 & 282.67 & 992 & 4 & 91.4T \\
9.5:0.5 & 3 & 313.0 & 0.319 & 48.44 & 4.83 & 279.31 & 280.75 & 1021 & 4 & 82.4T \\
9.5:0.5 & 4 & 554.3 & 0.180 & 65.71 & 35.00 & 163.67 & 516.15 & 1943 & 3 & 146.4T \\ \midrule
9:1 & 0 & 583.4 & 0.171 & 96.53 & 36.17 & 545.53 & 548.36 & 1954 & 8 & 60.2T \\
9:1 & 1 & 563.9 & 0.177 & 69.61 & 5.89 & 306.73 & 518.08 & 1947 & 8 & 31.9T \\
9:1 & 2 & 560.3 & 0.178 & 104.06 & 54.82 & 522.03 & 523.92 & 1959 & 8 & 148.4T \\
9:1 & 3 & 585.9 & 0.171 & 95.23 & 12.52 & 533.86 & 538.51 & 1949 & 8 & 51.1T \\
9:1 & 4 & 571.4 & 0.175 & 76.15 & 38.49 & 297.79 & 534.95 & 1943 & 8 & 145.3T \\ \bottomrule
\end{tabular}
\end{table*}
\begin{table*}[h]
\centering
\caption{Pipeline performance under scheduling, batching, buffering, and defense configurations. The unit of throughput is input per second; the unit of average end-to-end latency (AvgE2E), p50, p95, p99 is second; LPR\# denotes the amount of workload processed by the license-plate-recognition module; Drops denotes the number of workload dropped by bounded buffering.}
\label{tab:production}
\begin{tabular}{llrrrrrrrrr}
\toprule 

 \textbf{Config} & \textbf{Wall Time(s)} & \textbf{Throughput} & \textbf{AvgE2E} & \textbf{p50} & \textbf{p95} & \textbf{p99} & \textbf{LPR\#} & \textbf{Drops} & \textbf{Total FLOPs} \\ \midrule

 Clean & 17.3 & 0.578  & 1.23 & 0.49 & 3.86 & 4.68 & 6 & 0 & 3.27T \\
Attacked & 1805 & 0.006  & 503.4 & 634.6 & 900.8 & 916.6 & 9,315 & 0 & 215.3T \\
 conf.5 & 1084 & 0.009  & 373.1 & 678.3 & 859.2 & 863.4 & 5,576 & 0 & 129.6T \\
   b16 + conf.5 & 550 & 0.018  & 188.1 & 391.1 & 478.6 & 484.4 & 5,576 & 0 & 129.6T \\
  b16 + buf100 & 28.3 & 0.354  & 8.0 & 8.3 & 26.9 & 27.2 & 304 & 9,011 & 8.96T \\
  conf.5 + buf100 & 34.3 & 0.291  & 15.0 & 32.9 & 35.9 & 36.8 & 215 & 5,361 & 6.87T \\
 b16 + conf.5 + buf100 & 25.3 & 0.396  & 7.7 & 12.1 & 22.1 & 22.7 & 199 & 5,377 & 6.50T \\ \midrule
 Clean (buf100) & 18.2 & 0.550  & 1.25 & 0.49 & 3.85 & 4.65 & 6 & 0 & 3.27T \\ \bottomrule

\end{tabular}
\end{table*}

\textbf{Setup.}
This experiment tests whether the path-selection vulnerability persists under production-realistic deployment. We use the trained Traffic Monitoring variant of Figure~\ref{fig:traffic_monitoring_var}, which replaces the simulated pipeline's stand-ins with deployment-grade trained weights and uses bounded queues for inter-module communication, following production inference systems~\cite{hu2021scrooge,shen2019nexus,zhao2026pard}. We evaluate three defense categories. \emph{System-level configurations} (batched inference, bounded buffering, confidence-threshold filtering) are evaluated on the production pipeline under sustained adversarial traffic. \emph{Input-level preprocessing} (Gaussian noise, spatial smoothing) and \emph{SVM input filtering} (a classifier trained to detect adversarial inputs) are evaluated under mixed traffic at attack ratios of 10\%, 5\%, and 1\%, to test whether defense effectiveness depends on attack prevalence. Results appear in Tables~\ref{tab:clean-side-effect}, \ref{tab:attacked10}, \ref{tab:mix-study}, and~\ref{tab:svm-per-seed}.

\textbf{Without defenses, the production pipeline collapses to 1\% of clean throughput.}
Under sustained adversarial traffic, throughput drops from 0.578~img/s to 0.006~img/s, a 96$\times$ reduction. LPR workload rises from 6 to 9{,}315 invocations, and p99 latency reaches 916.6 seconds per image (Table~\ref{tab:production}). This is sustained operational state, not peak attack intensity. At 0.006~img/s the pipeline is unavailable for any deployment with real-time deadlines.

\textbf{Confidence thresholding and batching together provide negligible recovery.}
Raising the detection confidence threshold from 0.25 to 0.5 reduces LPR workload by 40\% (9{,}315 to 5{,}576), but throughput recovers only to 0.009~img/s. Adding batched inference (batch=16) doubles throughput to 0.018~img/s by amortizing per-call GPU overhead, but does not reduce total computation: the same 5{,}576 LPR invocations still execute, only faster per call. Without bounded buffering, the pipeline remains 32$\times$ below clean throughput across every combination of these defenses.

\textbf{Bounded buffering recovers throughput by sacrificing detection coverage.}
Adding bounded inter-module queues that drop items when full, a standard production pattern, restores throughput to 0.291--0.396~img/s, 50--69\% of clean. The recovery comes entirely from dropping 96.1--96.7\% of detection requests at the buffer boundary. Under clean inputs with the same buffer configuration, zero items are dropped (Table~\ref{tab:production}, last row); the drops under attack are entirely attacker-induced. The buffer cannot distinguish adversarial false positives from legitimate detections before discarding them, so in mixed deployment the same drop rate applies to every item in the queue. A traffic-monitoring system in this configuration would fail to record 96\% of the vehicles passing its sensors during an attack window, while reporting nominal throughput for the 4\% it processes. The attack is not absorbed by buffering; it is converted from compute exhaustion into coverage loss. The operator's dashboard shows recovery while the system has lost most of its detection capability.

\textbf{Input-level preprocessing reduces the gain of the attack but does not eliminate its mechanism.}
Gaussian noise ($\sigma=5/255$) and spatial smoothing (median filter, kernel=3) preserve throughput within 5\% of the no-defense baseline on clean inputs (Table~\ref{tab:clean-side-effect}). Under mixed traffic, both recover throughput in proportion to attack ratio. At 10\% attacked, preprocessing improves throughput from 0.037~img/s to 0.16--0.18~img/s, a 4--5$\times$ recovery that remains $\sim$3$\times$ below clean. At 1\% attacked, preprocessing reaches 0.52--0.58~img/s, near the clean baseline of 0.66, but LPR workload remains at 242--364 invocations versus 64 under clean, 4--6$\times$ above baseline. Preprocessing rounds off the perturbation magnitudes that produce path-vulnerable detections but does not eliminate them. Enough adversarial detections survive to keep LPR workload at multiples of clean. The mitigation operates one layer below where the attack's leverage lies; it reduces the gain of the attack rather than disabling its mechanism.

\textbf{SVM input filters fail bimodally at low attack ratios and degrade to no-defense at high ratios.}
A support vector machine (SVM) trained on 900 clean and 90 attacked images exhibits high seed-to-seed variance at low attack ratios. At 1\% attacked, throughput across five seeds is bimodal: three seeds catch the attack and reach 0.66--0.67~img/s, two miss and drop to 0.30~img/s (Table~\ref{tab:svm-per-seed}). The mean (0.523 $\pm$ 0.200~img/s) hides the structure: each seed samples a different attack subset, and the SVM either catches the subset cleanly or misses it entirely. At 100\% attacked (Table~\ref{tab:attacked10}), the filter degrades to 0.019~img/s, comparable to no defense, because no learned boundary can separate clean from adversarial when every input is adversarial. The filter is effective only when the deployed attack distribution matches its training distribution, a condition no defender can guarantee against an adaptive adversary.

\begin{tcolorbox}[breakable]
    \textbf{Answer to RQ4.}
The path-selection vulnerability persists in every defense configuration tested. No combination restores the production pipeline to clean operation under non-trivial attack ratios. The failure mode varies with the defense (compute exhaustion when scheduling is unconstrained, coverage loss with 96\% of detections dropped under buffering, 4--6$\times$ residual LPR workload under input preprocessing, distribution-dependent variance under ML-based filtering), but the vulnerability itself does not. Each defense category operates at the wrong layer: image-level preprocessing rounds off perturbation magnitudes without changing where the workload is routed; system-level scheduling bounds the rate at which workload reaches downstream modules without changing which modules are targeted; ML-based filters depend on attack-distribution assumptions that adaptive adversaries can violate. The attack lives at the routing layer between modules, where no defense in this taxonomy has visibility. Mitigating it will require routing-aware mechanisms (per-path workload budgets, anomaly detection on path-utilization distributions, or trust-bounded routing decisions) that current pipeline frameworks do not expose.
\end{tcolorbox}

\section{Discussion}
\label{sec:discussion}
 
\subsection{Parameter Estimation Challenges}
\label{sec:param_estimation}
 
Phase~1 ranking depends on three quantities: per-inference cost $c_v$, output cardinality $o_v$, and gating behavior $g_v$. The white-box adversary measures all three through instrumentation and code analysis. The gray-box adversary observes $c_v$ from published latency and FLOPs counts; $o_v$ and $g_v$ are not individually observable and collapse into the aggregate $g_v(o_v)$, inferred from input--output behavior under probe inputs.
 
The ranking depends on relative ordering of paths, not exact cost values. In every evaluated pipeline, the most expensive path exceeds the cheapest by more than an order of magnitude. In the production-realistic Traffic variant, the LPR path is roughly 32$\times$ more expensive than the FR path. For standard architectures, the precision of public model specifications sits well inside the margin needed to recover the correct ranking. Phase~1 reduces in practice to a public-information lookup. Against custom architectures with no published cost profile, $c_v$ must be estimated empirically; knowledge-agnostic estimation (timing side-channels, queue-occupancy traces) is left to future work.
 
\subsection{On Pipeline Realism}
\label{sec:realism_debate}
 
The five pipelines in Section~\ref{sec:whitebox} use publicly available architectures with untrained or partially trained weights. This design supports study of diverse pipeline topologies (single-branch, multi-branch, autoregressive- and classifier-terminated) without per-application training infrastructure. Conclusions depending on model accuracy, rather than on architectural cost structure, would be vulnerable on these pipelines.
 
The production-realistic Traffic variant in Section~\ref{sec:production} closes this gap. Every neural component is deployment-grade (RT-DETR-R50, MTCNN, InceptionResnetV1, DeepLabV3-ResNet50, GIT-base), and the inter-module substrate uses bounded queues, batched inference, and confidence thresholding from documented production systems~\cite{hu2021scrooge,shen2019nexus,zhao2026pard}. We do not claim the variant is itself deployed; we claim it reproduces the operational properties under which a deployed system would receive our adversarial inputs. The 96$\times$ throughput collapse without defenses, and 96.7\% silent data loss under bounded buffering, on this configuration cannot be dismissed as a simulation artifact.
 
\subsection{Limitations of the Gray-Box Threat Model}
\label{sec:graybox_limits}
 
Our gray-box adversary uses public architectural information (FLOPs, parameter counts, output schemas) for Phase~1 ranking, and same-family surrogate gradients for Phase~2 perturbation. The first capability is well-grounded: production systems use widely recognized architectures, and architecture choice often leaks through latency profiles, deployment documentation, or model fingerprinting. The second is standard in transfer-attack literature but requires a sufficiently close surrogate.
 
Two limitations follow. Against custom or proprietary architectures, Phase~1 falls back on empirical probing (Section~\ref{sec:param_estimation}). Against deployments without a transferable surrogate, Phase~2 weakens to query-based optimization, feasible but slower and noisier than gradient transfer. A fully strict black-box variant would require inferring $c_v$ and $g_v(o_v)$ from purely observational data and a perturbation step independent of surrogate gradients. We leave both to future work. The path-selection problem is independent of the threat model and becomes harder as adversary knowledge shrinks; the current evaluation establishes the upper end of that difficulty curve.
 
\subsection{Toward Routing-Aware Defenses}
\label{sec:routing_defenses}
 
The four defense categories evaluated in Section~\ref{sec:production} share a structural blind spot: none observes or constrains how workload is routed between modules. Effective mitigation requires routing-layer primitives. We see three. \textbf{Per-path workload budgets} cap activation rates relative to clean-traffic share, converting compute exhaustion into a visible quota miss rather than a silent throughput collapse. \textbf{Distributional anomaly detection on path utilization} aggregates evidence across many inputs, making sustained spoofing harder than for per-input filters like SVM. \textbf{Trust-bounded routing} propagates provenance metadata and gates expensive-path entry on accumulated trust, bounding per-input cost regardless of false-positive volume. None of these primitives is exposed by current pipeline frameworks. The work is not algorithmically deep; it requires treating per-path workload as a first-class operational quantity rather than a byproduct of model outputs.
\section{Related Work}
\label{sec:bg_attack}

Adversarial efficiency attacks represent an emerging threat class that targets the computational characteristics of deep neural networks rather than their classification accuracy. Unlike traditional adversarial examples that aim to induce misclassification, efficiency attacks preserve the correctness of model outputs while significantly degrading system performance.

We categorize prior work on efficiency attacks based on the specific dynamic behavior they exploit:

\noindent \textbf{Attacks on Per-Iteration Computational Cost.} These attacks increase the computational burden during each inference iteration while maintaining correct outputs. For image inputs, adversaries employ $L_2$ and $L_{\infty}$ norm-based perturbations to maximize computational expense by manipulating pixel values. For textual data, character and word-level perturbations compromise efficiency through imperceptible modifications~\cite{haque2020ilfo, chen2022deepperform, hong2020panda, pan2022gradauto, pan2023gradmdm}. These attacks primarily target dynamic behaviors where computational pathways vary based on input characteristics, such as conditional execution branches and early-exit mechanisms.
    
\noindent \textbf{Attacks on Inference Iteration Count.} These attacks manipulate neural networks to require additional iterations to generate final outputs, thereby increasing computational demands and reducing efficiency~\cite{chen2022nicgslowdown, chen2022nmtsloth, haque2023slothspeech, shumailov2021sponge}. Such attacks are particularly effective against autoregressive models, where output generation depends on sequential processing steps, and against models with adaptive refinement mechanisms that terminate based on convergence criteria or confidence thresholds.
    
\noindent \textbf{Attacks on Output Cardinality.} These attacks aim to increase the quantity of outputs generated by deep neural networks, amplifying the computational load on downstream tasks~\cite{shapira2023phantom, chen2024overload, ma2024slowtrack, liu2023slowlidar}. By manipulating the detection confidence of objects or features, adversaries can force models to generate excessive outputs that propagate through the pipeline, creating cascading performance degradation. These attacks are particularly effective against detection and segmentation models that feed into multi-stage processing pipelines.

While these attacks have been studied individually, our work presents the first systematic framework for analyzing and defending against them in the context of complete pipeline systems rather than isolated models. By considering the end-to-end impact of these attacks across interconnected components, we provide a more comprehensive understanding of system vulnerabilities and mitigation strategies.
\section{Conclusion}
\label{sec:conclusion}
 
Machine Learning (ML) inference pipelines composed of multiple models exhibit a vulnerability single-model attacks cannot exercise: the choice of which downstream branch a perturbation drives workload into. \tool\ formalizes this as path ranking followed by path-targeted perturbation. The ranking step alone accounts for amplification gaps of 1.5--2.9$\times$ over reduced variants and 3--117$\times$ over single-model baselines. Absolute amplification reaches 2{,}407$\times$ FLOPs and 263$\times$ latency on Expressway Surveillance.
 
The vulnerability is structural. It survives gray-box transfer (3.6--58.3$\times$ FLOPs with public model specifications and same-family surrogates). Every defense category evaluated under production deployment also fails: confidence thresholding, batching, bounded buffering, input preprocessing, and ML-based filtering each operate above the routing layer where the attack lives.
 
Effective mitigation requires routing-layer observability and policy that current pipeline frameworks do not expose. Path selection is a first-class adversarial problem; treating per-path workload as a first-class operational quantity is the corresponding defender's task.
\section{Ethics Considerations}

This work studies adversarial efficiency attacks on deep learning pipeline systems with the goal of exposing systemic vulnerabilities and informing defenses. All experiments ran in isolated, non-production environments under our control, against models we hosted ourselves; no live systems, third-party APIs, or end users were targeted, and no real-world workloads were degraded in the course of this study. The pipelines were assembled from publicly released models and standard benchmark datasets used in accordance with their licenses. Released code is intended solely for academic reproduction and defense research, and we omit configuration details that would substantially shorten the path to a turn-key attack on production systems. We acknowledge the dual-use nature of efficiency-degradation techniques, but consider open investigation of these risks essential to building trustworthy AI systems, and we hope our findings inform routing-aware defenses in future pipeline frameworks.

\bibliographystyle{IEEEtran}
\bibliography{NDSS_BIB}

\clearpage

\appendices

\section{Pipeline Applications}
\label{sec:pipeline_applications}

As shown in Figure~\ref{fig:five_applications}, we developed five distinct pipeline applications. These applications were chosen to present a trade-off between diversity and real-world fidelity, evaluating different combinations of neural and non-neural components. The details of the five pipeline applications are as follows:

\textbf{Traffic Monitoring. }The traffic monitoring pipeline processes an input stream by first routing it through a YOLO-based object detection module. The detected bounding boxes are then processed by two parallel downstream components: an Inception ResNet model for person recognition and a CRNN model for license plate recognition. Finally, the extracted data is aggregated by a non-neural summary module to produce the final output.

\textbf{Wildlife Conservation Area Surveillance. }This application monitors and documents the status of protected animal habitats. From the initial input , a YOLO object detection module identifies coarse animal categories in the frame. This is followed by a ResNet-50-based fine-grained animal classification model  that distinguishes specific individuals within a species. A non-neural summary component then compiles this information into the final output.

\textbf{Smart Home Appliance Alert System. }Designed to monitor residential safety , this pipeline takes environmental input and utilizes YOLO for object detection to identify household appliances. A ResNet-50-based hazard detection module then assesses the operational state of these devices to flag potential risks. To conclude the workflow, a Large Language Model (Qwen2-72B-Instruct) processes the detected states to generate a comprehensive safety summary as the final output.

\textbf{Expressway Vehicle Surveillance. }This pipeline aims to monitor traffic violations on expressways. The input stream is analyzed by a YOLO object detection module to isolate vehicles. The isolated vehicle data is then processed simultaneously by a CRNN module for license plate recognition and a ResNet-50-based model for violation detection  (e.g., unlawful driving behavior). A non-neural summary module consolidates these findings to generate the final output.

\textbf{Amber Alert System. }This system is designed to detect dangerous situations and automatically generate AMBER alerts. Following the visual input , a YOLO object detection module locates persons in the scene. The pipeline then splits into three specialized neural components: person recognition using InceptionResNet , age estimation using a Vision Transformer (ViT) model , and post/pose estimation using an additional ViT model. These aggregated insights are passed to an LLM (Qwen2-70B-Instruct), which formats the data to generate the final alert message for dispatch.

\end{document}